\documentclass{article}
\usepackage{PRIMEarxiv}
\usepackage[utf8]{inputenc}
\usepackage[T1]{fontenc}
\usepackage{graphicx}
\usepackage{multirow}
\usepackage{amsmath,amssymb,amsfonts}
\usepackage{amsthm}
\usepackage{mathrsfs}
\usepackage[title]{appendix}
\usepackage{xcolor}
\usepackage{textcomp}
\usepackage{booktabs}
\usepackage{algorithm}
\usepackage{algorithmicx}
\usepackage{algpseudocode}
\usepackage{subcaption}
\usepackage{hyperref}
\usepackage{fancyhdr}

\graphicspath{{figures/}}

\begin{document}

\title{A Stability-Aware Frozen Euler Autoencoder for Physics-Informed Tracking in Continuum Mechanics (SAFE-PIT-CM)}

\author{
  Emil Hovad \\
  Section for Statistics and Data Analysis, DTU Compute \\
  Technical University of Denmark, Kgs.\ Lyngby, Denmark \\
  \texttt{emilh@dtu.dk}
}

\maketitle

\begin{abstract}
Material parameters such as thermal diffusivity govern 
how microstructural fields evolve during processing, 
but difficult to measure directly.
The Stability-Aware Frozen Euler
Physics-Informed Tracking for Continuum Mechanics
(SAFE-PIT-CM), is an autoencoder that
embeds a frozen convolutional layer as a differentiable
PDE solver in its
latent-space transition to jointly recover diffusion
coefficients and the underlying physical field from
temporal observations.
When temporal snapshots are saved at intervals coarser
than the simulation time step, a single forward Euler
step violates the von~Neumann stability condition,
forcing the learned coefficient to collapse to an
unphysical value.
Sub-stepping with SAFE restores stability at negligible
cost---each sub-step is a single frozen convolution,
far cheaper than processing more frames---with recovery error converging
monotonically with substep count.
Validated on thermal diffusion in metals, the method
recovers both the diffusion coefficient and the
physical field with near-perfect accuracy, both
with and yet without pre-training.
Backpropagation through the frozen operator supervises
an attention-based parameter estimator without labelled
data. The architecture generalises to any PDE with a
convolutional finite-difference discretisation.
\end{abstract}


\section{Introduction}
\label{sec:intro}

Diffusion coefficients govern how fields evolve in time and ultimately
determine the properties of many types of engineering challenges.
Recovering these coefficients and the governing field values from 
observational data are essential for calibrating computational models, 
yet classical approaches such as manual fitting, genetic algorithms, 
and gradient-free optimisation scale poorly with the number of unknowns.
Physics-informed neural networks (PINNs) \cite{raissi2019physics} embed
the governing PDE directly into the loss function and can identify
parameters from data. PINNs, however, require pointwise access to the
solution field---position, time, and the corresponding field value
(e.g.\ pressure, temperature, or concentration) at each collocation
point---and are poorly suited to video-like observational data where the
mapping between pixel intensities and physical quantities is unknown.
A different strategy embeds differentiable physics simulators directly
inside neural network architectures.
De Avila Belbute-Peres et al.\ \cite{de2018end} demonstrated this for
rigid-body dynamics, achieving end-to-end learning of physical parameters
by differentiating through a linear complementarity problem (LCP) solver.
Their work showed that structured physics knowledge, embedded as a frozen
differentiable layer, yields dramatically better sample efficiency than
black-box alternatives. Their approach, however, targets rigid-body
mechanics rather than continuum PDEs, and the numerical challenges
differ fundamentally: rigid-body LCP solvers do not face the von~Neumann
stability constraints inherent to explicit time integration of parabolic
equations.
Neural operators \cite{li2021fourier} and graph-based simulators
\cite{sanchez2020learning} learn dynamics from data but do not recover
material parameters directly. Solver-in-the-loop approaches
\cite{um2020solver} combine learned corrections with traditional solvers
but target simulation accuracy rather than parameter identification.
Physics-informed machine learning more broadly
\cite{karniadakis2021physics, brunton2020machine} has demonstrated the
value of combining physical knowledge with learned components, yet a gap
remains: \emph{no existing method provides a physics-informed autoencoder
whose latent-space transition is a frozen, stability-aware continuum PDE
solver for joint recovery of diffusion coefficients and the underlying physical field from temporal data.}
This gap matters for explainability. Black-box video models can fit data
accurately but offer no physical insight into \emph{why} a field evolves
as it does. An architecture whose latent dynamics are governed by a known
PDE is inherently interpretable---the learned diffusion
coefficient~$\alpha$ carries direct physical meaning, and the latent
propagation can be inspected step by step.
This paper presents \textbf{SAFE-PIT-CM}
(\textbf{S}tability-\textbf{A}ware \textbf{F}rozen \textbf{E}uler
autoencoder for \textbf{P}hysics-\textbf{I}nformed \textbf{T}racking in
\textbf{C}ontinuum \textbf{M}echanics), extending the
differentiable-physics-as-a-layer paradigm from rigid-body dynamics to
continuum PDEs.  The architecture is an autoencoder in which the
latent-space transition is governed by a frozen PDE operator rather than
a learned layer: a convolutional encoder maps video frames to a latent
field; the SAFE operator propagates the latent field forward in time via
sub-stepped finite differences; and a decoder reconstructs the video.
Because the physics is embedded as a frozen, differentiable layer,
backpropagation through the SAFE operator yields gradients that directly
supervise an attention-based estimator for the diffusion coefficient.

Like the differentiable physics engine of \cite{de2018end}, SAFE-PIT-CM
embeds a frozen physics solver inside a neural network and
backpropagates through it to learn physical parameters. The critical
difference is \emph{numerical stability}. When temporal data is saved at
intervals coarser than the original simulation time step, a forward
Euler step at the frame-to-frame interval violates the von~Neumann
stability condition, causing the learned diffusion coefficient to
collapse to an unphysical value. The SAFE operator resolves this by
decomposing each frame-to-frame transition into multiple small Euler
steps that respect the stability bound. This sub-stepping is
\emph{essential} for correct parameter recovery---a failure mode that,
to the best of the author's knowledge, has not been identified in prior
work on differentiable PDE solvers.
Crucially, sub-stepping is also computationally cheap: each sub-step is
a single frozen convolution, making it far less expensive than the
alternative of saving frames at finer temporal resolution, which would
require regenerating and storing additional training data.

The main contributions are:
\begin{enumerate}
    \item \textbf{SAFE-PIT-CM}: an autoencoder architecture whose
          latent-space transition is a frozen PDE operator rather than a
          learned layer, enabling unsupervised diffusion coefficient recovery
          from video data using only the physics residual as supervision.
    \item The \textbf{SAFE operator}: a frozen, fully differentiable
          finite-difference PDE operator with sub-stepping that enforces
          numerical stability, enabling gradient-based parameter
          optimisation through the solver.
    \item An \textbf{attention-based parameter estimator} that infers
          per-simulation diffusion coefficients from the temporal structure
          of the field, recovering $\alpha$ without ground-truth labels.
    \item \textbf{Test-time training (TTT)}: the ability to learn $\alpha$ from
          a \emph{single} simulation with no pre-training, using only the
          SAFE loss as supervision---analogous to the test-time optimisation
          phase of PINO~\cite{li2024physics}.
    \item \textbf{Computational efficiency}: sub-stepping with frozen
          convolutions restores numerical stability at negligible cost,
          avoiding the need to process more frames in the network.
\end{enumerate}

SAFE-PIT-CM is validated on the diffusion equation, jointly recovering
the diffusion coefficient and the temporal field across the physically
relevant range for engineering metals.
The architecture is lightweight, trains in minutes on a single GPU, and
test-time training converges in seconds per simulation.
The modular design generalises to any PDE admitting a convolutional
finite-difference discretisation, making it broadly applicable to
inverse problems in continuum mechanics.

\section{Method}
\label{sec:method}

\subsection{Governing equation}

The governing equation is the diffusion equation on a two-dimensional periodic
domain:
\begin{equation}
\frac{\partial u}{\partial t} = D \, \nabla^2 u, \qquad D > 0
\label{eq:diffusion}
\end{equation}
where $u(\mathbf{x}, t) \in \mathbb{R}$ is a scalar field (e.g., temperature)
and $D > 0$ is the diffusion coefficient.
Diffusion is a smoothing process: spatial gradients decrease over time.

The non-dimensional diffusion coefficient is
$\alpha^{*} = D\,\tau / \mathcal{L}^{2}$, where $\mathcal{L}$ and
$\tau$ are the characteristic length and time scales.
As an example, thermal diffusivities of common engineering metals span roughly
$D \approx 3 \times 10^{-6}$\,m$^2$/s (stainless steel~304) to
$D \approx 1.7 \times 10^{-4}$\,m$^2$/s (silver)~\cite{incropera2006fundamentals},
corresponding to $\alpha^{*} \in [0.03,\, 1.7]$ for a 10\,mm sample
observed over 1\,s. The training range $\alpha^{*} \in [0.01,\, 1.7]$ covers this full span.
This formulation captures the essential numerical structure of any
parabolic PDE whose dominant spatial operator is the Laplacian. The
governing equation is deliberately kept simple to isolate the key
contribution (the SAFE operator) from application-specific complexity.

The Laplacian is discretized on a uniform grid with spacing
$\Delta x = \Delta y$ using the standard five-point stencil with
periodic boundary conditions:
\begin{equation}
\nabla^2 u_{i,j} \approx \frac{u_{i+1,j} + u_{i-1,j} - 4\,u_{i,j} + u_{i,j+1} + u_{i,j-1}}{\Delta x^2}
\label{eq:laplacian}
\end{equation}
Time integration uses explicit (forward) Euler:
$u^{n+1} = u^n + \Delta t_{\text{sim}} \, \alpha \, \nabla^2 u^n$.

\paragraph{Training data.}
The training set comprises 100 synthetic simulations ($T=101$ frames each,
$\Delta t_{\text{save}} = 0.25$) where each video frame $V_t$ is the
scalar field $u(\cdot, t)$ itself, stored as a single-channel image.
The model therefore operates directly on the physical field rather than
on a rendered visualisation of it.
Full details and data parameters are given in
Appendix~\ref{app:data-generation} and Table~\ref{tab:data-params}.

\subsection{The SAFE-PIT-CM architecture}

\begin{figure}[t]
  \centering
  \includegraphics[width=\textwidth]{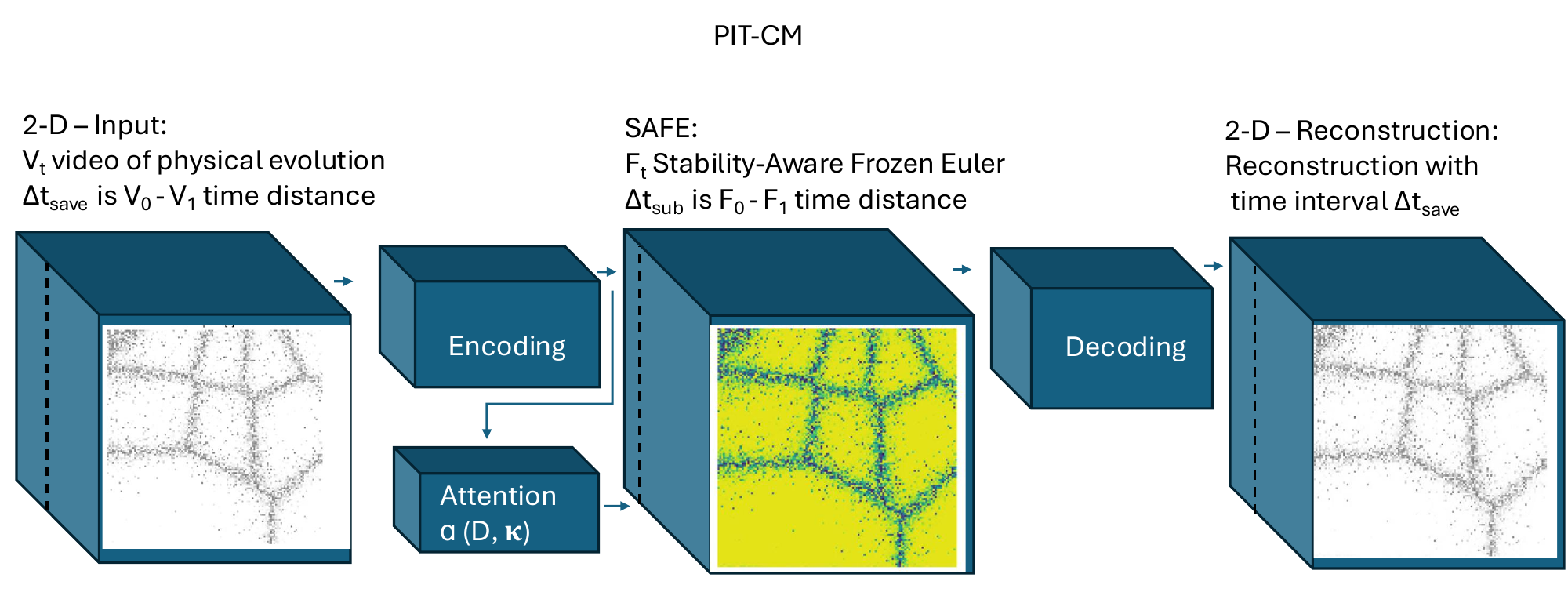}
  \caption{The SAFE-PIT-CM architecture. Input frames $V_t$ are separated
  by $\Delta t_{\text{save}}$. The encoder maps each frame to a latent
  field $F_t$; the SAFE operator propagates $F_t$ forward using
  $N_{\text{sub}}$ sub-steps of size
  $\Delta t_{\text{sub}} = \Delta t_{\text{save}} / N_{\text{sub}}$
  to predict $\hat{F}_{t+1}$; the decoder reconstructs $\hat{V}_t$
  from $F_t$. The diffusion coefficient $\hat{\alpha}$ is inferred by
  an attention-based estimator from the temporal structure of $\{F_t\}$.
  Dashed lines indicate the time intervals $\Delta t_{\text{save}}$
  and $\Delta t_{\text{sub}}$.}
  \label{fig:architecture}
\end{figure}

The SAFE-PIT-CM model (Fig.~\ref{fig:architecture}) is an autoencoder that
processes a single-channel video of $T$ frames,
$\{V_t\}_{t=0}^{T-1}$ with $V_t \in \mathbb{R}^{1 \times H \times W}$,
through four stages.
For each consecutive pair of frames $(t, t\!+\!1)$, the encoder maps both
frames to latent fields $F_t$ and $F_{t+1}$; the SAFE operator takes $F_t$
and produces a prediction $\hat{F}_{t+1}$ by applying the finite-difference
PDE operator $N_{\text{sub}}$ times (sub-steps); and the SAFE loss penalises
$\|\hat{F}_{t+1} - F_{t+1}\|^2$.
Over $T$ input frames this yields $T\!-\!1$ physics-residual comparisons
that supervise the diffusion coefficient $\hat{\alpha}$ without labels.
The architecture is shown in Fig.~\ref{fig:architecture} and Table~\ref{tab:architecture}.

\begin{table}[t]
\centering
\caption{Network architecture.  All convolutions use $3\!\times\!3$ kernels
with circular padding (periodic boundaries) unless noted.
The SAFE operator is frozen (no trainable parameters).}
\label{tab:architecture}
\begin{tabular}{llr}
\toprule
\textbf{Component} & \textbf{Description} & \textbf{Parameters} \\
\midrule
Encoder        & 3 conv blocks + $1\!\times\!1$ proj.; zero-init
               & ${\sim}7\text{k}$ \\
Physics est.   & 2 conv + attention ($d\!=\!32$, 4 heads) + MLP
               & ${\sim}9\text{k}$ \\
SAFE operator  & $N_{\text{sub}}$ frozen Laplacian conv steps
               & $9$ (frozen) \\
Decoder        & 2 conv blocks + $1\!\times\!1$ proj.; zero-init
               & ${\sim}5\text{k}$ \\
\midrule
\textbf{Total} & & ${\sim}\textbf{21k}$ \\
\bottomrule
\end{tabular}
\end{table}

\paragraph{Encoder.}
A ResNet-style convolutional block maps each video frame into a latent field:
$F_t = V_t + g_\theta(V_t)$, where $g_\theta$ consists of three convolutional
blocks ($1 \to c \to c \to c$ channels, with batch normalization, ReLU
activations, circular padding for periodic boundaries, and skip connections)
followed by a $1 \times 1$ projection back to one channel. Crucially, the
projection weights are \textbf{initialized to zero}, so that $F_t = V_t$ at the
start of training. This prevents the encoder from immediately distorting the
spatial structure of the field, which would destroy the Laplacian signal
needed for parameter estimation.

\paragraph{Physics estimator.}
An attention-based module estimates the diffusion coefficient $\alpha$ from the full
temporal sequence of latent fields. Each frame $F_t \in \mathbb{R}^{1 \times H \times W}$
is spatially compressed by adaptive average pooling to $4 \times 4$,
followed by two convolutional layers ($1 \to d/2 \to d$ channels, where
$d = d_{\text{model}} = 32$) and a final global average pool, yielding a
per-frame feature vector $\mathbf{z}_t \in \mathbb{R}^d$. The sequence
$(\mathbf{z}_0, \ldots, \mathbf{z}_{T-1})$ is processed by multi-head
self-attention \cite{vaswani2017attention} with $n_{\text{heads}} = 4$,
following the factorised spatial-then-temporal design of video
transformers~\cite{arnab2021vivit, bertasius2021timesformer},
a residual connection, and layer normalization. Mean pooling across the time
dimension produces a single simulation-level representation, from which
a two-layer MLP head with Softplus activation predicts
$\hat{\alpha}$ directly.

By attending across all time steps simultaneously, the estimator
has access to the global temporal evolution of the field, enabling
it to compare rates of spatial change across the full sequence.

\paragraph{Frozen PDE operator (SAFE operator).}
A non-trainable module implements Eq.~\ref{eq:diffusion} with forward Euler
time integration. The Laplacian (Eq.~\ref{eq:laplacian}) is computed as a
$3 \times 3$ convolution with circular padding and frozen kernel weights
$[0, 1, 0; \; 1, -4, 1; \; 0, 1, 0] / (\Delta x \, \Delta y)$. Given one
latent field $F_t$ as an example and the estimated $\hat{\alpha}$, the
operator produces the predicted next frame $\hat{F}_{t+1}$.
In its simplest form ($N_{\text{sub}} = 1$, a single step):
\begin{equation}
\hat{F}_{t+1} = F_t + \Delta t_{\text{save}} \cdot \hat{\alpha} \cdot \nabla^2 F_t
\label{eq:nsub-one}
\end{equation}
However, this formulation is \emph{numerically unstable} for
most parameter values of interest (Section~\ref{sec:substepping}).

\paragraph{Decoder.}
A lightweight decoder (2 conv blocks) maps the latent field back to video space:
$\hat{V}_t = F_t + h_\phi(F_t)$, with the same zero-init projection ensuring
$\hat{V}_t = F_t$ initially.

\subsection{The SAFE operator: stability-aware sub-stepping}
\label{sec:substepping}

This is the central technical contribution. The training data consists of
field snapshots saved every $s$ simulation steps, giving a frame-to-frame
interval $\Delta t_{\text{save}} = s \cdot \Delta t_{\text{sim}}$. For the present data,
$s = 50$ and $\Delta t_{\text{sim}} = 0.005$, so $\Delta t_{\text{save}} = 0.25$.

The forward Euler scheme applied to the discrete diffusion equation is
stable if and only if the von~Neumann condition is satisfied:
\begin{equation}
\Delta t_{\text{sub}} \cdot \alpha \cdot \lambda_{\max} < 2
\label{eq:stability}
\end{equation}
where $\lambda_{\max}$ is the spectral radius of the five-point Laplacian
stencil. On a uniform periodic grid with $\Delta x = \Delta y$,
$\lambda_{\max} = 8 / \Delta x^2$ (Appendix~\ref{app:von-neumann}); for $\Delta x = 0.5$ this gives
$\lambda_{\max} = 32$. When $N_{\text{sub}} = 1$ (Eq.~\ref{eq:nsub-one}),
the operator uses $\Delta t_{\text{sub}} = \Delta t_{\text{save}} = 0.25$, giving the stability
requirement:
\begin{equation}
0.25 \cdot \alpha \cdot 32 < 2 \quad \Rightarrow \quad \alpha < 0.25
\label{eq:stability-violation}
\end{equation}
Since the data spans $\alpha \in [0.01, 1.7]$, \emph{most simulations
violate the stability condition}. During training, the optimizer cannot
increase $\alpha$ above ${\sim}0.25$ without causing the forward pass to
diverge; it therefore converges to a small, stable value that minimizes
the physics residual within the stable region, typically
$\hat{\alpha} \approx 0.02$, an order of magnitude below the true values.

\paragraph{Solution: the SAFE operator.}
The SAFE operator decomposes each frame-to-frame transition into
$N_{\text{sub}}$ sub-steps of size
$\Delta t_{\text{sub}} = \Delta t_{\text{save}} / N_{\text{sub}}$.
For every source field $F_t$, $t = 0, \ldots, T\!-\!2$, the operator
iterates:
\begin{equation}
F_t^{(0)} = F_t, \qquad
F_t^{(k+1)} = F_t^{(k)} + \Delta t_{\text{sub}} \cdot \hat{\alpha} \cdot \nabla^2 F_t^{(k)}, \qquad
k = 0, \ldots, N_{\text{sub}} - 1
\label{eq:substep}
\end{equation}
producing the predictions
$\hat{F}_{[1,\,\ldots,\,T-1]}$, where
$\hat{F}_{t+1} := F_t^{(N_{\text{sub}})}$.
After $N_{\text{sub}}$ sub-steps, each prediction has been advanced by
exactly $\Delta t_{\text{save}} = N_{\text{sub}} \cdot \Delta t_{\text{sub}}$.
Setting $N_{\text{sub}} = s = 50$
recovers the original simulation time step $\Delta t_{\text{sub}} = 0.005$,
and the stability condition becomes:
\begin{equation}
0.005 \cdot \alpha \cdot 32 = 0.16 \, \alpha < 2 \quad \Rightarrow \quad \alpha < 12.5
\label{eq:stability-safe}
\end{equation}
which is satisfied for any physically reasonable diffusion coefficient.
Table~\ref{tab:timescales} summarises the three time scales and
Table~\ref{tab:stability} (Sect.~\ref{sec:ablation}) shows the stability properties and recovery quality for each tested $N_{\text{sub}}$ value.

\begin{table}[h]
\centering
\caption{Time scales in SAFE-PIT-CM.  Two distinct time steps govern
the method: $\Delta t_{\text{save}}$, the interval \emph{between saved frames}
in the input data, and $\Delta t_{\text{sub}}$, the internal time step used
by the SAFE operator for sub-stepping \emph{within} each frame pair.
The sub-step count $N_{\text{sub}}$ is a tuneable hyperparameter; the default is
$N_{\text{sub}} = 50$ (matching the save interval $s$), ablated over
$N_{\text{sub}} \in \{2, 5, 10, 25, 50\}$ in
Sect.~\ref{sec:ablation}}
\label{tab:timescales}
\begin{tabular}{lllll}
\toprule
\textbf{Symbol} & \textbf{Name} & \textbf{Where} & \textbf{Definition} & \textbf{Value} \\
\midrule
$\Delta t_{\text{sim}}$   & Simulation time step & Data generation          & (PDE solver step)                         & $0.005$ \\
$s$                       & Save interval        & Data generation          & (steps between frames)                    & $50$ \\
\midrule
$\Delta t_{\text{save}}$  & Frame interval       & Between frames           & $s \cdot \Delta t_{\text{sim}}$           & $0.25$ \\
\midrule
$N_{\text{sub}}$          & Sub-step count       & SAFE operator            & (hyperparameter)                          & $50$ (default) \\
$\Delta t_{\text{sub}}$   & Sub-step size        & Within SAFE operator     & $\Delta t_{\text{save}} / N_{\text{sub}}$ & $0.005$ (at $N_{\text{sub}}\!=\!50$) \\
\bottomrule
\end{tabular}
\end{table}


Because each sub-step $F_t^{(k)} \to F_t^{(k+1)}$ is a differentiable
operation (frozen convolution + scalar multiplication + addition), gradients
propagate cleanly through all $N_{\text{sub}}$ iterations via
backpropagation through time, reaching the parameter estimator with a
well-conditioned signal.

\paragraph{Implications for differentiable physics.}
Previous work on differentiable physics engines \cite{de2018end} operates in
the rigid-body regime, where the LCP formulation does not involve explicit
time integration and von~Neumann stability constraints do not arise. In the
continuum PDE setting, numerical stability of the explicit integrator is
an unavoidable concern. Any architecture that embeds a frozen explicit PDE
solver as a differentiable layer must account for the stability of the time
integrator, or risk learning parameters dictated by numerics rather than
physics. This stability--accuracy coupling has not been addressed in prior
work on differentiable PDE solvers.

\subsection{Loss function}

\begin{figure}[t]
  \centering
  \includegraphics[width=\textwidth]{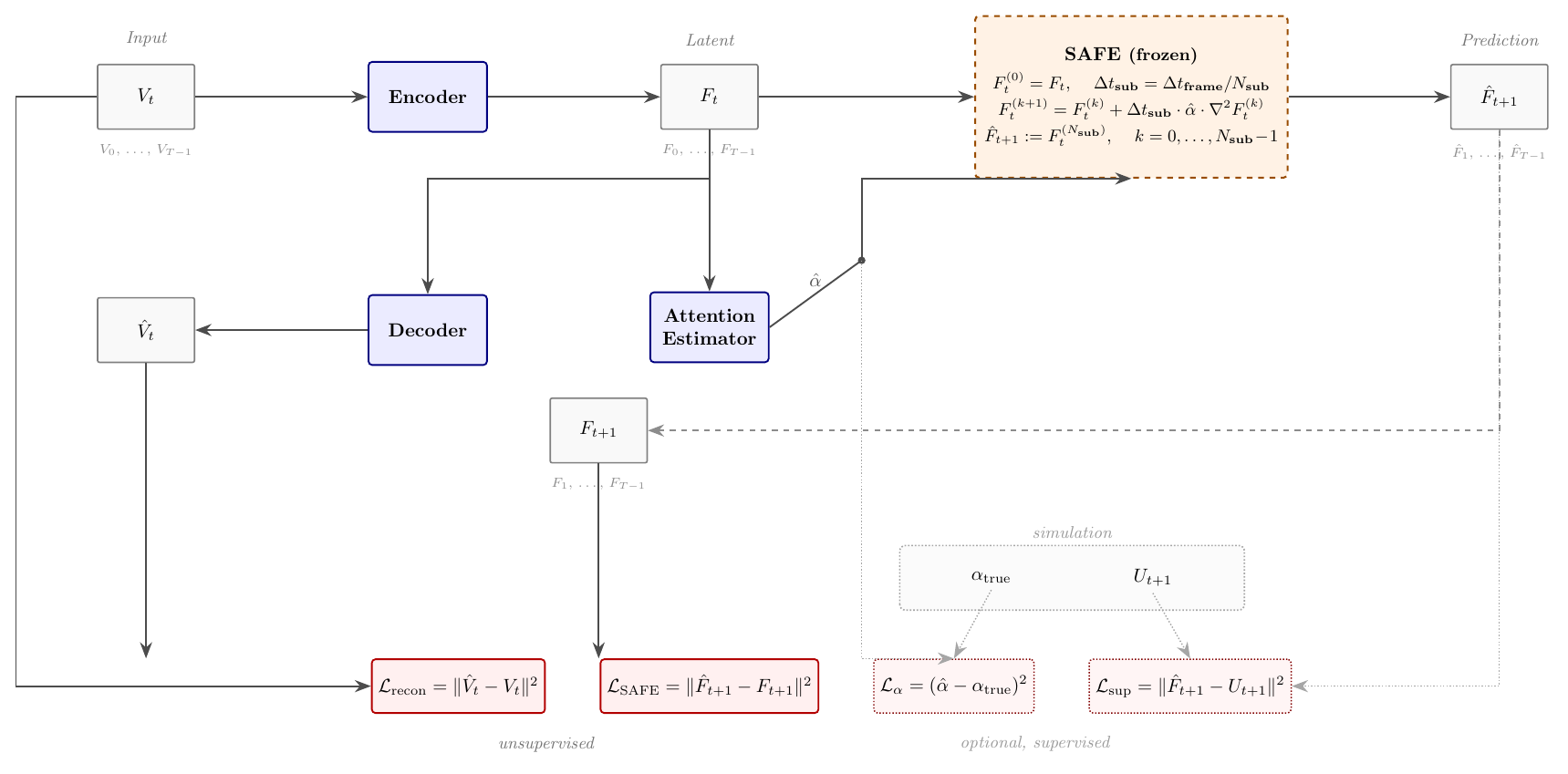}
  \caption{Data flow and training losses. The encoder maps each frame $V_t$
  to a latent field $F_t$; the attention-based estimator predicts
  $\hat{\alpha}$ from the sequence $\{F_t\}$; the frozen SAFE operator
  iterates $N_{\text{sub}}$ sub-steps (Eq.~\ref{eq:substep}) to produce
  $\hat{F}_{t+1}$; the decoder reconstructs $\hat{V}_t$ from $F_t$.
  Solid red boxes: unsupervised losses; dotted red boxes: optional supervised
  losses available when simulation data is provided}
  \label{fig:arch-flow}
\end{figure}

The model is trained with two core losses
(see Fig.~\ref{fig:arch-flow}):
\begin{equation}
\mathcal{L} = w_{\text{SAFE}} \, \mathcal{L}_{\text{SAFE}}
            + w_{\text{recon}} \, \mathcal{L}_{\text{recon}}
\label{eq:loss}
\end{equation}

\paragraph{SAFE loss (physics residual).}
The SAFE operator (Eq.~\ref{eq:substep}) advances each encoded field
$F_t$ by one frame interval to produce $\hat{F}_{t+1}$.
The SAFE loss penalises the mismatch between the prediction and the
independently encoded next frame:
\begin{equation}
\mathcal{L}_{\text{SAFE}}
  = \frac{1}{T\!-\!1} \sum_{t=0}^{T-2}
    \bigl\|\,\hat{F}_{t+1} - F_{t+1}\bigr\|^2
\label{eq:safe-loss}
\end{equation}
Because both sides depend on the learnable encoder and the estimated
$\hat{\alpha}$, backpropagation through the frozen SAFE operator
provides direct gradient supervision for
$\hat{\alpha}$---no ground-truth labels are required.

\paragraph{Reconstruction loss (data fidelity).}
The reconstruction loss ensures the encoder--decoder pair preserves
the input video:
\begin{equation}
\mathcal{L}_{\text{recon}} = \frac{1}{T} \sum_{t=0}^{T-1} \left\| \hat{V}_t - V_t \right\|^2
\end{equation}

\paragraph{Optional supervised extensions.}
When ground-truth labels are available, the method can also operate
in a supervised setting by adding
$\mathcal{L}_{\alpha} = (\hat{\alpha} - \alpha_{\text{true}})^2$
to the loss. A soft bounds penalty
$\mathcal{L}_{\text{bounds}}$ can further constrain $\hat{\alpha}$
to a physically plausible interval.

\subsection{Training}
\label{sec:training}

All learnable parameters---encoder, decoder, and physics estimator---are
optimised jointly with both the SAFE loss and the reconstruction loss.
The only component that is never updated is the SAFE operator itself:
the Laplacian stencil is a fixed physical law, not a learned approximation.

A single model is trained on the full $\alpha$ range without requiring
separate models or process-type flags.

The zero-init projection in the encoder ($F_t = V_t$ at initialisation)
provides a natural bootstrap: on the first forward pass the physics estimator
sees the raw field, giving an immediate and unambiguous gradient signal for
$\alpha$.  As training proceeds the encoder learns a refined latent
representation while the reconstruction loss prevents it from collapsing to a
trivial flat field.

\subsection{Test-time training (TTT)}
\label{sec:zero-shot}

SAFE-PIT-CM can recover $\alpha$ from a \emph{single simulation} without
any pre-training, following the test-time optimisation paradigm of
PINO~\cite{li2024physics}. Given one video sequence, a fresh (randomly
initialised) model is optimised on the SAFE loss alone, with the encoder
and decoder frozen at their zero-init state ($F_t = V_t$).
Only the SAFE loss is needed because the zero-init skip connection
makes the reconstruction already exact ($\hat{V}_t = V_t$) before any
training begins; the reconstruction loss therefore carries no gradient
signal, and the SAFE loss provides the \emph{sole} learning signal for
$\hat{\alpha}$.

Because the SAFE operator provides a strong inductive bias---relating
$\hat{\alpha}$ directly to frame-to-frame smoothing---the parameter
estimator converges to the correct $\alpha$ within
${\sim}100$--$500$ gradient steps, which corresponds to a few seconds of
wall-clock time on a single GPU.
Convergence-based stopping (relative tolerance $10^{-6}$, patience
50~steps, cap 2\,000~steps) ensures that no unnecessary computation
is performed.
This is orders of magnitude faster than genetic-algorithm or grid-search
calibration and does not require any training data.

TTT also provides \emph{generalisation without retraining}: a single
simulation at any $\alpha$ value---even outside the training range---can
be processed, since the model starts fresh each time and adapts to the
data purely through the physics loss.
The term ``test-time training'' emphasises \emph{when} learning occurs
(at inference), while ``zero-shot'' emphasises \emph{what} is needed
beforehand (nothing).

\section{Implementation details}
\label{sec:implementation}

SAFE-PIT-CM is implemented in PyTorch as a pip-installable Python package.
Table~\ref{tab:hyperparams} summarises the training hyperparameters.
The encoder uses $c = 16$ intermediate channels. The physics
estimator uses $d_{\text{model}} = 32$ with 4 attention heads.

\begin{table}[h]
\centering
\caption{Training hyperparameters}
\label{tab:hyperparams}
\begin{tabular}{lll}
\toprule
\textbf{Parameter} & \textbf{Symbol} & \textbf{Value} \\
\midrule
Optimizer          &                                 & Adam \\
Learning rate      & $\eta$                         & $1 \times 10^{-3}$ \\
Scheduler          &                                 & ReduceLROnPlateau (0.5, pat.\ 5) \\
Early stopping     &                                 & patience 5 \\
SAFE loss weight   & $w_{\text{SAFE}}$              & 100 \\
Recon.\ loss weight & $w_{\text{recon}}$            & 1 \\
Sub-steps          & $N_{\text{sub}}$               & 50 (1, 2, 5, 10, 25 in conv.\ analysis) \\
Frame interval      & $\Delta t_{\text{save}}$       & 0.25 \\
SAFE sub-step size & $\Delta t_{\text{sub}}$        & 0.005 \\
Batch size         & $B$                            & 4 \\
Epochs             &                                 & 50 \\
\bottomrule
\end{tabular}
\end{table}

During test-time inference, optimisation runs until convergence (relative tolerance $10^{-6}$,
patience 50~steps, cap 2\,000~steps).
For trained-model inference (\emph{pretrained mode}), each held-out simulation
is fine-tuned until convergence: all learnable parameters are updated jointly.
A single model is trained on the full $\alpha$ range (see Sect.~\ref{sec:training}).
For test-time training (TTT), a fresh model is optimised until convergence per
simulation using the SAFE loss alone, with the encoder and decoder kept at
their zero-init state ($F_t = V_t$) so the physics estimator receives a
direct signal from the raw data.

\section{Results}
\label{sec:results}

\begin{figure}[t]
  \centering
  \begin{subfigure}[b]{0.48\textwidth}
    \centering
    \includegraphics[width=\textwidth]{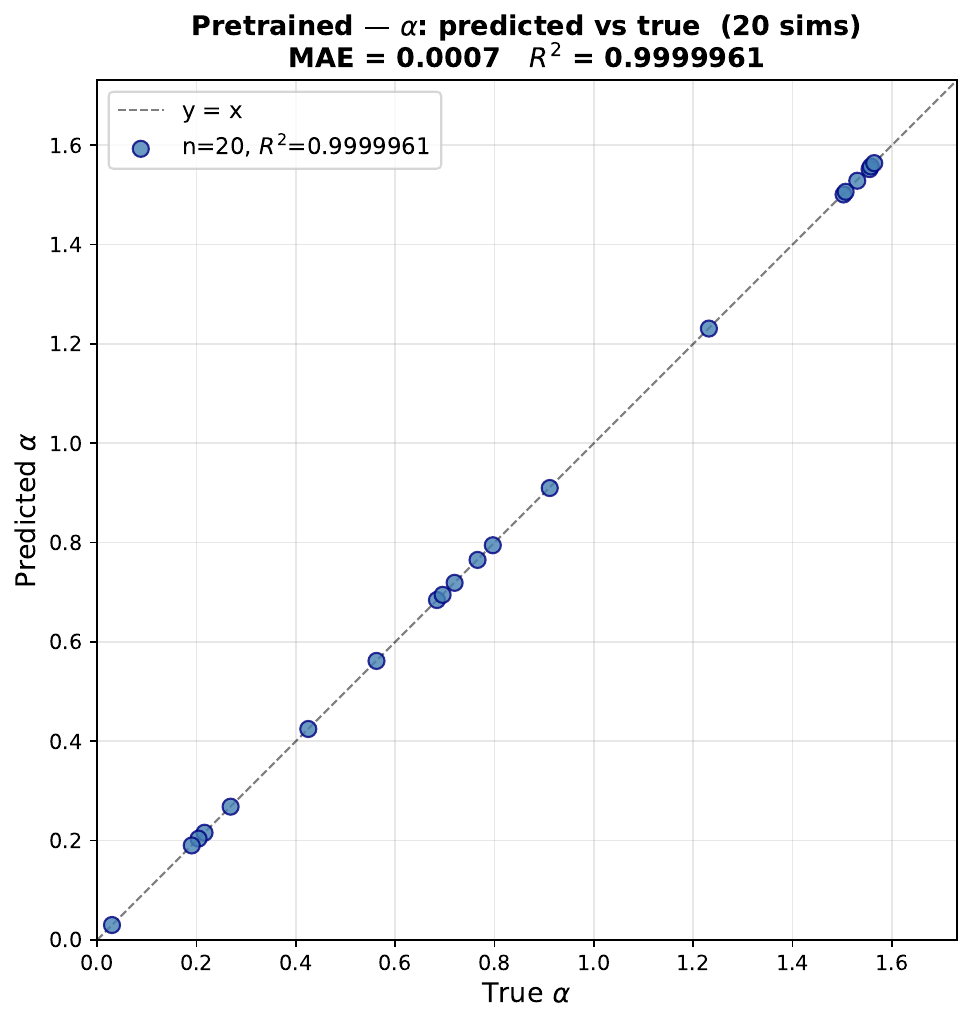}
    \caption{Inference after training (tested on held-out simulations).}
    \label{fig:alpha-scatter-trained}
  \end{subfigure}
  \hfill
  \begin{subfigure}[b]{0.48\textwidth}
    \centering
    \includegraphics[width=\textwidth]{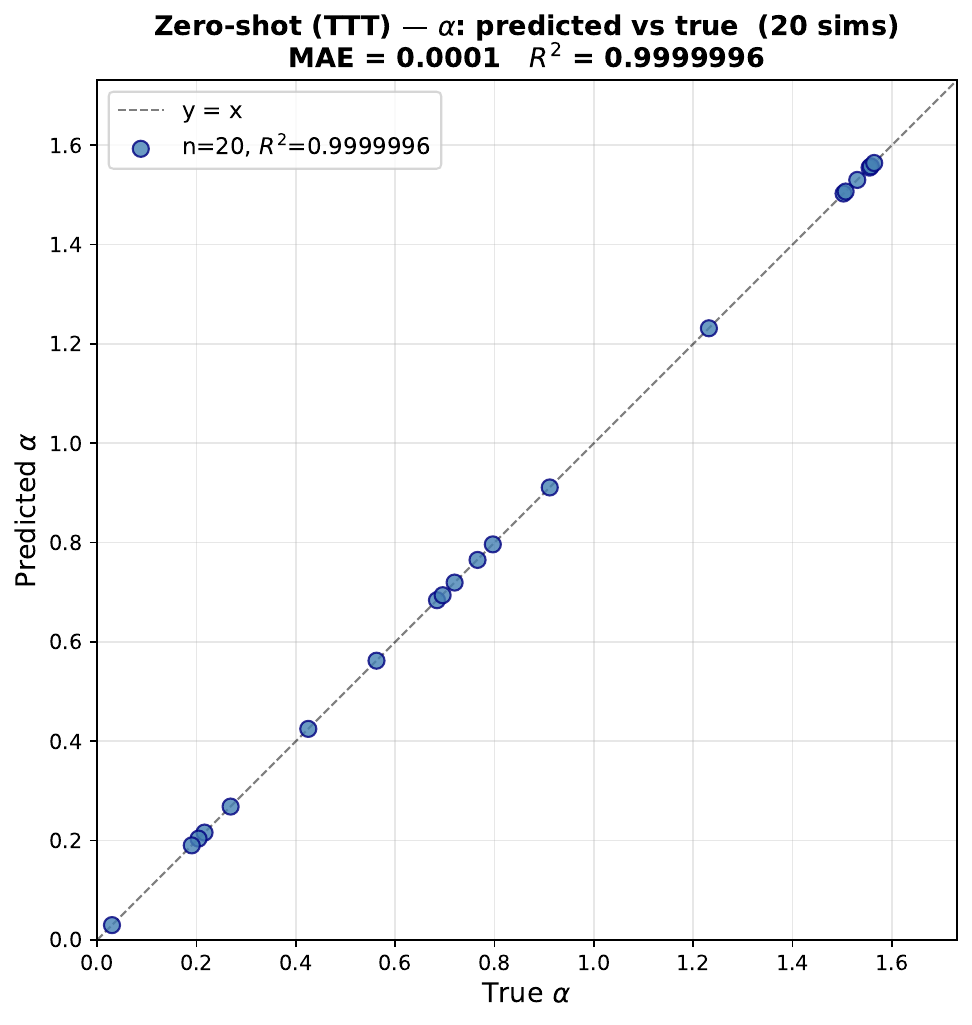}
    \caption{Test-time training (no pre-training, fresh model per simulation).}
    \label{fig:alpha-scatter-zeroshot}
  \end{subfigure}
  \caption{Predicted vs.\ true diffusion coefficient $\alpha$ on held-out
  simulations.
  (a)~Pretrained inference: model fine-tuned
  until convergence per test simulation with $\mathcal{L}_{\text{SAFE}} + \mathcal{L}_{\text{recon}}$
  (all learnable parameters updated; SAFE operator frozen).
  (b)~Test-time training: fresh model optimised until convergence per simulation
  with $\mathcal{L}_{\text{SAFE}}$ alone (no pre-training required)}
  \label{fig:alpha-scatter}
\end{figure}

\paragraph{Parameter recovery.}
Table~\ref{tab:results} and Fig.~\ref{fig:alpha-scatter} summarise the
quantitative results on 20 held-out simulations.
Pure inference --- running the pretrained model with a single forward
pass and no test-time adaptation --- yields MAE$\,=\,0.67$ and
$R^2 = -1.55$, confirming that the pretrained model alone cannot recover
$\alpha$ without per-simulation adaptation.
In pretrained + TTT (Fig.~\ref{fig:alpha-scatter-trained}), the model is
fine-tuned per simulation until convergence with all learnable parameters
updated jointly, achieving MAE$\,=\,4.1 \times 10^{-4}$ and
$R^2 = 0.9999985$.
In zero-shot TTT (Fig.~\ref{fig:alpha-scatter-zeroshot}), a fresh model
is optimised per simulation with the SAFE loss alone, requiring no
pre-training, and achieves MAE$\,=\,2.9 \times 10^{-5}$ and
$R^2 = 1.0000000$.
Table~\ref{tab:field-errors} reports the physical field recovery quality.
\begin{table}[h]
\centering
\caption{Diffusion coefficient recovery on 20 held-out simulations,
$\alpha \in [0.03,\, 1.56]$}
\label{tab:results}
\begin{tabular}{lcc}
\toprule
\textbf{Setting} & \textbf{MAE}$_\alpha$ & $\boldsymbol{R^2}$ \\
\midrule
Pure inference   & $6.7 \times 10^{-1}$ & $-1.547$ \\
Pretrained + TTT & $4.1 \times 10^{-4}$ & $1.000$ \\
Zero-shot TTT    & $2.9 \times 10^{-5}$ & $1.000$ \\
\bottomrule
\end{tabular}
\end{table}

\begin{table}[h]
\centering
\caption{Physical field and physics residual quality ($N_{\text{sub}} = 50$).
The SAFE loss $\mathcal{L}_{\text{SAFE}}$ (Eq.~\ref{eq:safe-loss})
measures the physics residual in latent space;
$\mathcal{L}_{\text{recon}} = \|\hat{V}_t - V_t\|^2$ measures the
physical field recovery error, since the video frames are direct
observations of the field.  Both are mean squared errors averaged
over all frames and simulations}
\label{tab:field-errors}
\begin{tabular}{lcccc}
\toprule
\textbf{Setting} & \textbf{$n$} & \textbf{Mean $\mathcal{L}_{\text{SAFE}}$} & \textbf{Mean $\mathcal{L}_{\text{recon}}$} \\
\midrule
Pure inference   & 20 & $1.5 \times 10^{-6}$ & $5.3 \times 10^{-5}$ \\
Pretrained + TTT & 20 & $4.4 \times 10^{-10}$ & $1.7 \times 10^{-7}$ \\
Zero-shot TTT    & 20 & $8.1 \times 10^{-11}$ & $2.5 \times 10^{-7}$ \\
\bottomrule
\end{tabular}
\end{table}

\paragraph{Sub-stepping is essential.}
The sub-step ablation (Fig.~\ref{fig:substep-ablation} and
Table~\ref{tab:stability}) confirms that numerical stability is a
prerequisite for accurate parameter recovery.
At $N_{\text{sub}} \leq 2$, the von~Neumann stability condition
(Eq.~\ref{eq:stability-violation}) is violated for most of the data
range, and recovery collapses entirely.
At $N_{\text{sub}} = 10$ the stability bound is satisfied for all
$\alpha \in [0.01, 1.7]$, and performance improves monotonically up to
$N_{\text{sub}} = 50$, which matches the original simulation save
interval and reduces the per-step stability number by a factor of~50.

\paragraph{Test-time training.}
The TTT mode (Fig.~\ref{fig:alpha-scatter-zeroshot}) demonstrates
the strength of the SAFE operator as an inductive bias. With no
pre-training and no learned encoder, optimising the SAFE loss on a single
simulation recovers $\alpha$ with MAE\,$=$\,$2.9 \times 10^{-5}$ and $R^2 = 1.0000000$.
The Laplacian operator
provides a direct, physics-grounded relationship between $\alpha$ and
the frame-to-frame field evolution: faster smoothing implies larger $\alpha$.

\begin{figure}[t]
  \centering
  \begin{subfigure}[b]{\textwidth}
    \centering
    \includegraphics[width=\textwidth]{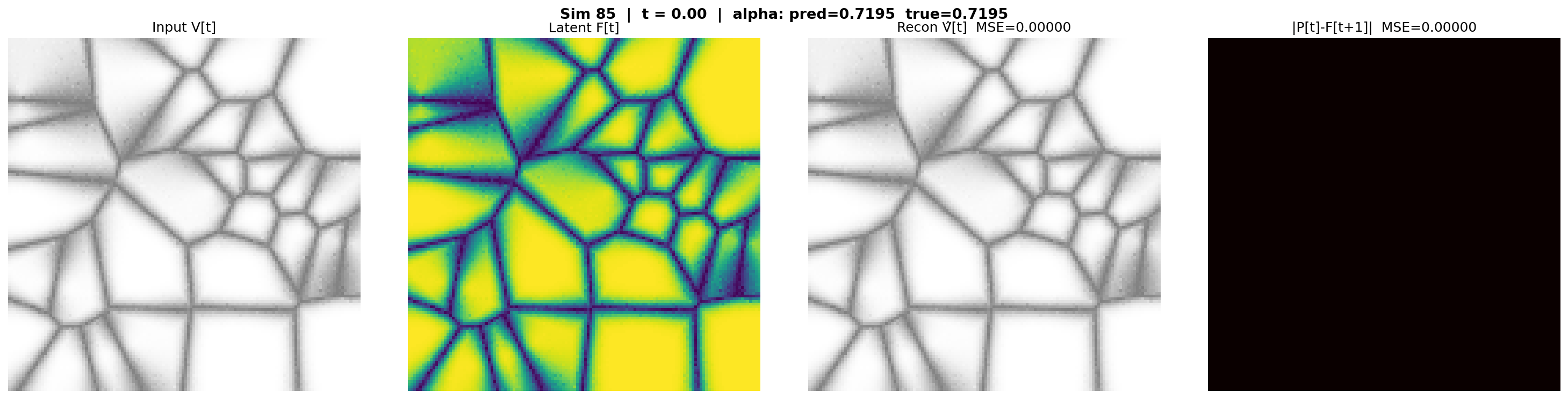}
    \caption{$t = 0$ (initial condition).}
    \label{fig:diff-t000}
  \end{subfigure}
  \vspace{0.3em}
  \begin{subfigure}[b]{\textwidth}
    \centering
    \includegraphics[width=\textwidth]{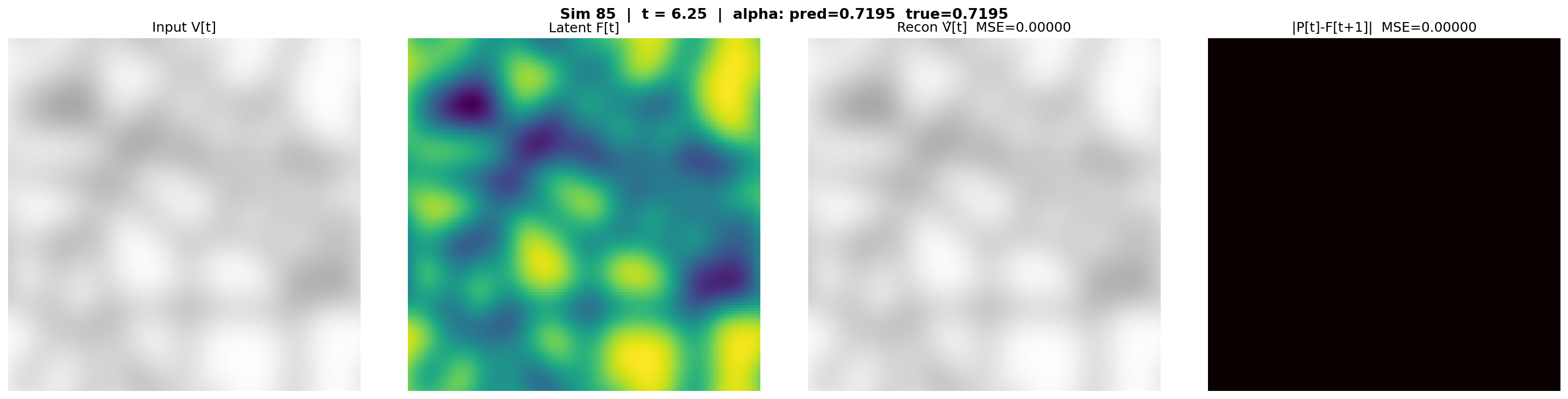}
    \caption{$t = 6.25$.}
    \label{fig:diff-t025}
  \end{subfigure}
  \vspace{0.3em}
  \begin{subfigure}[b]{\textwidth}
    \centering
    \includegraphics[width=\textwidth]{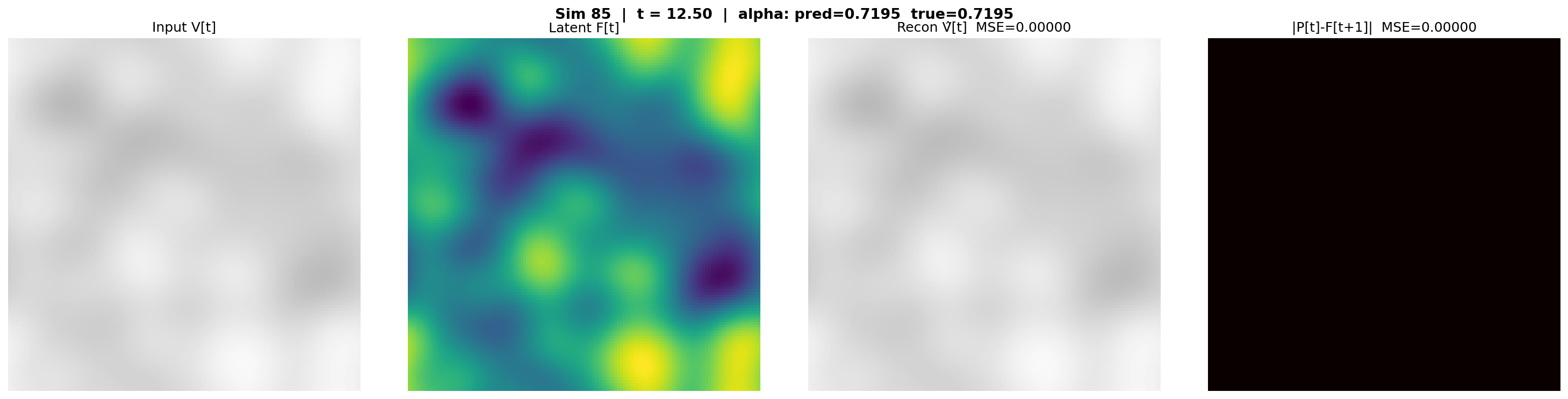}
    \caption{$t = 12.50$.}
    \label{fig:diff-t050}
  \end{subfigure}
  \vspace{0.3em}
  \begin{subfigure}[b]{\textwidth}
    \centering
    \includegraphics[width=\textwidth]{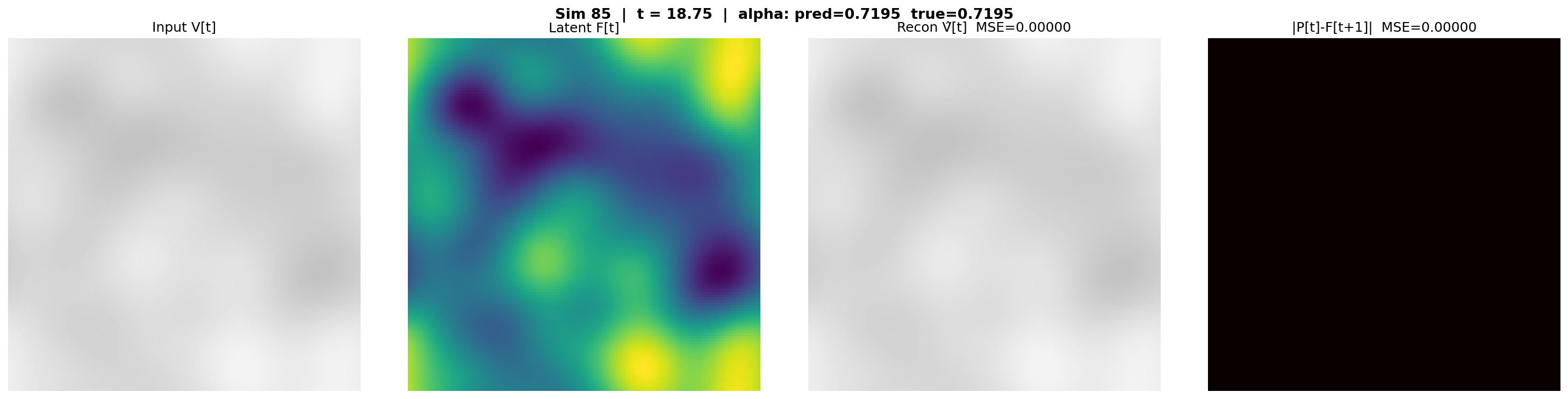}
    \caption{$t = 18.75$.}
    \label{fig:diff-t075}
  \end{subfigure}
  \caption{Example inference on a held-out simulation.
  Each row shows (left to right): input field $V_t$, latent field $F_t$,
  reconstructed field $\hat{V}_t$, and absolute error $|\hat{F}_{t+1} - F_{t+1}|$.
  The field evolves from a sharp Voronoi microstructure~(a) through
  progressive smoothing~(b,\,c) toward a diffused state~(d)}
  \label{fig:example-diffusion}
\end{figure}

\paragraph{Training dynamics.}
$\alpha$ converges rapidly (within ${\sim}5$--10 epochs).
The high SAFE loss weight ($w_{\text{SAFE}} = 100$) keeps the physics residual
dominant, while the reconstruction loss prevents the encoder from degenerating
to a flat latent field.

\subsection{Convergence analysis}
\label{sec:ablation}

Three ablation studies isolate the contributions of sub-stepping,
initialisation, and temporal resolution.

\begin{figure}[t]
  \centering
  \includegraphics[width=\textwidth]{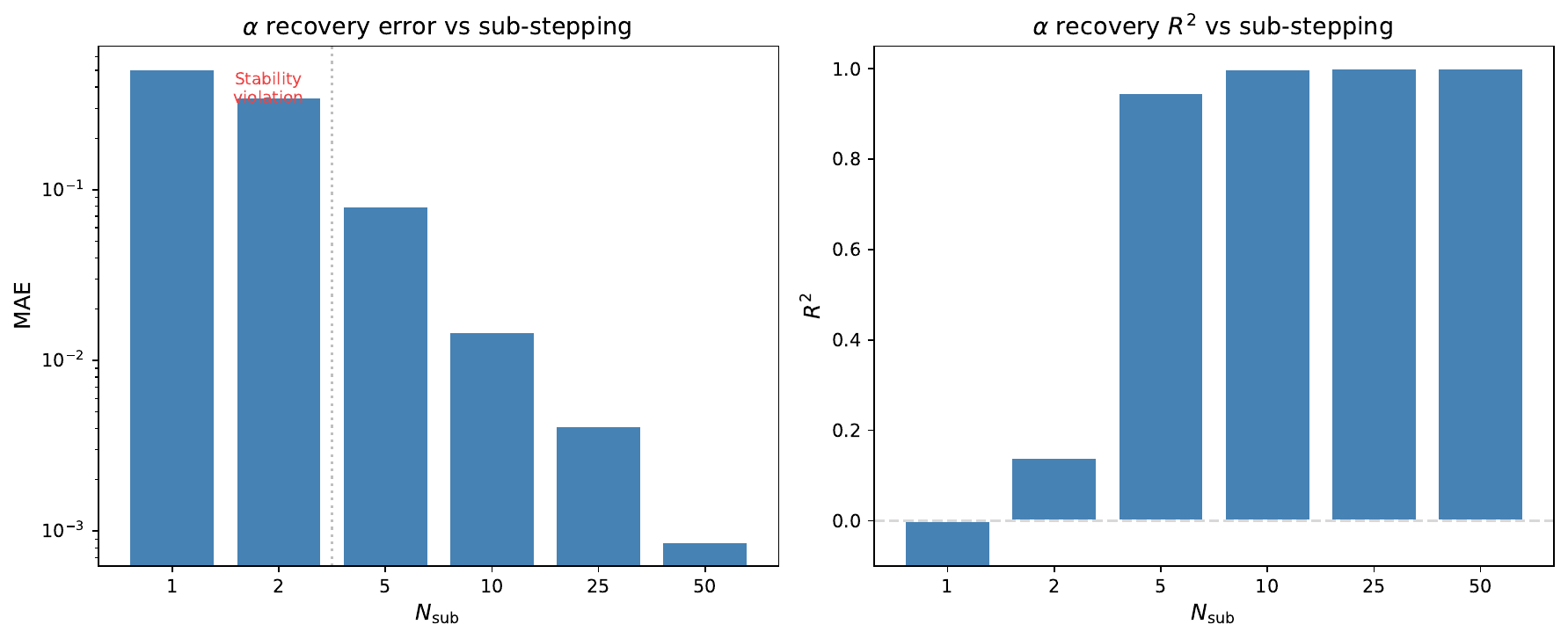}
  \caption{Convergence of $\alpha$ recovery with sub-step count $N_{\text{sub}}$ (zero-shot TTT). Left: MAE (log scale).
  Right: $R^2$. Recovery collapses at $N_{\text{sub}} \leq 2$
  due to the stability violation; performance converges
  monotonically with increasing $N_{\text{sub}}$}
  \label{fig:substep-ablation}
\end{figure}

\paragraph{Effect of $N_{\text{sub}}$ --- zero-shot.}
Fig.~\ref{fig:substep-ablation} shows $\alpha$ recovery MAE and $R^2$
as a function of $N_{\text{sub}} \in \{1, 2, 5, 10, 25, 50\}$ in zero-shot mode
(fresh model, SAFE loss only).
For $N_{\text{sub}} \leq 2$, the von~Neumann stability condition is
violated for large $\alpha$, causing recovery to collapse.
At $N_{\text{sub}} = 50$ (matching the simulation save interval),
recovery is near-perfect.

\begin{table}[h]
\centering
\caption{Stability properties and zero-shot $\alpha$ recovery for each
tested sub-step count $N_{\text{sub}}$.
The von~Neumann stability number
$\sigma = \Delta t_{\text{sub}} \cdot \alpha \cdot \lambda_{\max}$ must satisfy
$\sigma < 2$; increasing $N_{\text{sub}}$ reduces $\Delta t_{\text{sub}}$ and
expands the recoverable $\alpha$ range.  The data spans
$\alpha \in [0.01, 1.7]$; $\sigma_{\max}$ is evaluated at $\alpha = 1.7$.
The last three columns report zero-shot TTT recovery quality on 20 held-out simulations.
$\mathcal{L}_{\text{recon}}$ is zero to machine precision in all cases
because the encoder and decoder are frozen at their zero-init state
(see also Fig.~\ref{fig:substep-ablation})}
\label{tab:stability}
\begin{tabular}{cccccccc}
\toprule
$\boldsymbol{N_{\text{sub}}}$ & $\boldsymbol{\Delta t_{\text{sub}}}$ & $\boldsymbol{\sigma_{\max}}$ & $\boldsymbol{\alpha_{\max}}$ & \textbf{Stable?} & $\mathrm{MAE}_\alpha$ & $\boldsymbol{R^2}$ & $\mathcal{L}_{\text{SAFE}}$ \\
\midrule
1   & 0.2500 & 13.60 & 0.25  & No  & 0.5042 & $-0.5242680$ & $6.0 \times 10^{-6}$ \\
2   & 0.1250 & 6.80 & 0.50  & No  & 0.3449 & $0.1385034$ & $3.0 \times 10^{-6}$ \\
5   & 0.0500 & 2.72 & 1.25  & No  & 0.0798 & $0.9468073$ & $1.0 \times 10^{-7}$ \\
10  & 0.0250 & 1.36 & 2.50  & Yes & 0.0153 & $0.9987317$ & $1.0 \times 10^{-8}$ \\
25  & 0.0100 & 0.54 & 6.25  & Yes & 0.0048 & $0.9998469$ & $1.0 \times 10^{-8}$ \\
50  & 0.0050 & 0.27 & 12.50  & Yes & 0.0006 & $0.9999926$ & $3.0 \times 10^{-11}$ \\
\bottomrule
\end{tabular}
\end{table}

\paragraph{Effect of $N_{\text{sub}}$ --- pretrained.}
The same convergence pattern holds when starting from the pre-trained
checkpoint (pretrained + TTT), confirming that sub-stepping is essential
regardless of initialisation.

\section{Discussion and Conclusion}
\label{sec:conclusion}

SAFE-PIT-CM extends the differentiable-physics-as-a-layer paradigm
from rigid-body dynamics \cite{de2018end} to continuum PDEs.  Unlike
rigid-body solvers whose gradients can be computed analytically without
stability concerns, the explicit Euler integrator introduces a von~Neumann
stability constraint that, if ignored, silently corrupts the learned
parameters.  The SAFE operator resolves this by enforcing stability through
sub-stepping.
Unlike black-box video prediction models, the architecture is inherently
interpretable: every prediction traces to the encoder's field mapping,
a step-by-step Euler propagation with an inspectable $\hat{\alpha}$, and
the decoder reconstruction.  The predicted $\alpha$ is directly comparable
to values from independent experiments (e.g., dilatometry, calorimetry),
providing a built-in sanity check.
The approach is also related to PINNs \cite{raissi2019physics} and PINO
\cite{li2024physics}, which minimise the PDE residual at collocation points.
SAFE-PIT-CM differs by minimising the residual on the discrete grid using the
same finite-difference stencil as the data generator, making the operator
exact (modulo floating-point precision) at the true $\alpha$ and requiring no
pointwise field access.
The dominant computational cost is the SAFE operator:
$\mathcal{O}(N_{\text{sub}} \times H \times W)$ per frame pair, amounting to
50 convolutions per forward pass at $N_{\text{sub}} = 50$ on $128\times 128$
grids; training on 80 simulations completes in minutes on a single GPU.

\paragraph{Zero-shot TTT outperforms pre-training.}
Zero-shot TTT achieves the best results across all metrics
(Table~\ref{tab:results}). In the noise-free setting studied here, the
encoder converges to near-identity regardless of initialisation, meaning
pre-training offers no representational advantage. A pretrained encoder
introduces a small bias that the subsequent TTT must overcome, whereas
a fresh model optimises $\hat{\alpha}$ without this overhead. This
explains the consistently lower errors of zero-shot TTT and suggests
that pre-training becomes beneficial only when the encoder must learn a
non-trivial mapping, e.g.\ from noisy or indirect observations.

\paragraph{Well-posedness.}
Because diffusion is a smoothing process, the SAFE operator always runs
in the numerically stable forward direction. Every Fourier mode decays,
the physics loss at the true $\alpha$ is zero (with the identity encoder),
and there is no need for reverse-direction formulations or process-type
flags.

\paragraph{Diffusion coefficient and physical field recovery.}
SAFE-PIT-CM jointly recovers the diffusion coefficient $\alpha = D$ and
the underlying physical field from temporal data. Because diffusion is a
smoothing process---spatial gradients decrease over time---the SAFE
operator always runs in the numerically stable forward direction.

This paper presented SAFE-PIT-CM, an autoencoder that embeds a frozen,
stability-aware PDE solver as its latent-space transition, enabling
unsupervised recovery of diffusion coefficients and the
underlying physical field from temporal data.
The central contribution is the SAFE operator: a frozen, fully
differentiable finite-difference operator with sub-stepping that enforces
the von~Neumann stability condition. Without sub-stepping, the learned
diffusion coefficient collapses to an unphysical value dictated by
numerical stability rather than the data---a failure mode not previously
identified in the differentiable physics literature.
SAFE-PIT-CM recovers both the diffusion coefficient $\alpha = D$ and the
physical field with near-perfect accuracy ($R^2 > 0.999998$) across a range
spanning two orders of magnitude, and supports test-time training
($R^2 = 1.0000000$) from a single simulation with no pre-training.
The attention-based physics estimator attends across all time steps
simultaneously---including future frames---comparing the rate of spatial
change between early and late frames, a signal directly proportional
to~$\alpha$. This is fundamentally different from frame-by-frame
(recurrent) approaches, which see only local temporal gradients and
cannot exploit the global temporal structure.
The modular design extends to any PDE with a convolutional
finite-difference discretisation, opening a path toward automated
material parameter identification from video observations of physical
processes.

\paragraph{Limitations.}
As a method paper, the validation is performed on synthetic data with
noise-free field observations. In this setting the encoder converges to
near-identity, and parameter recovery is driven entirely by the SAFE
operator; the encoder--decoder pair is not required to learn a non-trivial
field representation. With noisy or indirect observations the encoder
would need to learn a denoising or inverse mapping, exercising the full
architecture. Extending SAFE-PIT-CM to such settings is left as future
work.

\section*{Acknowledgements}

This work was supported by the Villum Foundation under the Villum Experiment project ``Scientific machine learning for advancing the understanding of the 4-D microstructural evolution in metals (SciML4D)'' (grant no.\ 77978). The code and simulation data will be made publicly available upon acceptance.

\section*{Disclaimer}
The software and methods presented in this work are provided as-is, without warranty. The author assumes no liability for any use of the code or results.

\bibliographystyle{unsrt}
\bibliography{references}

\begin{appendices}

\section{Data generation}
\label{app:data-generation}

\begin{table}[h]
\centering
\caption{Data generation parameters}
\label{tab:data-params}
\begin{tabular}{lll}
\toprule
\textbf{Parameter} & \textbf{Symbol} & \textbf{Value} \\
\midrule
Grid size               & $N_x \times N_y$          & $128 \times 128$ \\
Grid spacing            & $\Delta x = \Delta y$     & $0.5$ \\
Time step               & $\Delta t_{\text{sim}}$   & $0.005$ \\
Number of time steps    & $N_{\text{step}}$         & $5000$ \\
Save interval           & $s$                       & $50$ \\
Saved frame interval    & $\Delta t_{\text{save}}$  & $0.25$ \\
Frames per simulation   & $T$                       & $101$ \\
Number of Voronoi cells & --                        & $15$--$35$ \\
Interface width         & $w$                       & $2.5$ \\
Diffusion coefficient   & $\alpha = D$              & $\mathcal{U}(0.01,\; 1.7)$ \\
Total simulations       & --                        & $100$ \\
Training / test split   & --                        & $80\%$ / $20\%$ \\
\bottomrule
\end{tabular}
\end{table}

All training data is generated by numerically integrating the diffusion
equation (Eq.~\ref{eq:diffusion}) on a $128 \times 128$ periodic grid
with $\Delta x = \Delta y = 0.5$, using forward Euler time integration
with $\Delta t_{\text{sim}} = 0.005$ for 5000 steps per simulation.
Snapshots are saved every 50 steps, yielding 101 frames per simulation
(including the initial condition at $t = 0$). The von~Neumann stability
condition (Eq.~\ref{eq:stability}) is verified before each run.

\subsection{Initial conditions}

The initial condition is a Voronoi tessellation of 15--35 cells on the
periodic domain. Each cell centre is placed uniformly at random. For
each grid point, the two nearest cell centres are identified, and the
scalar field is set via a smooth $\tanh$ profile of the distance
difference:
\begin{equation}
u_{i,j} = \tfrac{1}{2}\bigl(1 + \tanh\bigl((d_2 - d_1 + \xi) \,/\, w\bigr)\bigr)
\end{equation}
where $d_1$ and $d_2$ are the distances to the nearest and second-nearest
cell centres, $w = 2.5$ is the interface width, and $\xi$ is a spatially
varying perturbation that introduces position-dependent boundary roughness.
The composite field $u = \max_i(\eta_i)$ takes values near~1 in cell
interiors and ${\sim}0.5$ at cell boundaries.

\subsection{Simulation data}

100 simulations are generated by integrating
\begin{equation}
\frac{\partial u}{\partial t} = \alpha \, \nabla^2 u, \qquad \alpha \sim \mathcal{U}(0.01,\; 1.7)
\end{equation}
using forward Euler with $\Delta t_{\text{sim}} = 0.005$ for 5000 steps. The
von~Neumann stability condition requires
\begin{equation}
\Delta t_{\text{sim}} \cdot \alpha \cdot \lambda_{\max} < 2, \qquad
\lambda_{\max} = \frac{8}{\Delta x \, \Delta y} = 32
\end{equation}
giving $\alpha < 2 / (0.005 \times 32) = 12.5$, which is satisfied
for all sampled values. The simulation code verifies this condition
before each run and reduces $\Delta t_{\text{sim}}$ if needed.

Each simulation evolves from a sharp Voronoi microstructure toward a
progressively smoother field as boundaries diffuse outward.
Snapshots are saved every $s = 50$ steps, yielding 101 frames per
simulation with $\Delta t_{\text{save}} = s \cdot \Delta t_{\text{sim}} = 0.25$.
Each simulation models thermal diffusion with a different diffusion
coefficient.

\subsection{Data splits}

An 80/20 train/test split is applied, giving a representative mix of
$\alpha$ values in each set. Each simulation is stored as a single array
of shape $[T, N_x, N_y]$ with associated metadata (diffusion coefficient,
grid parameters, time step).

\section{Von Neumann stability analysis}
\label{app:von-neumann}

This appendix derives the Von Neumann stability condition for the
forward Euler (FTCS) scheme applied to the 2D diffusion equation
\begin{equation}
\frac{\partial u}{\partial t} = \alpha \, \nabla^2 u
\label{eq:app-diffusion}
\end{equation}
The analysis follows Morton \& Mayers~\cite{morton2005}
(Ch.~2--3, Theorem~5.2) and is standard; we include it for completeness
and to make the connection to the SAFE operator explicit.

\subsection{Two spatial dimensions}
\label{app:vn-2d}

The 2D diffusion equation $u_t = \alpha\,(u_{xx} + u_{yy})$ is discretised
using the five-point stencil on a grid with spacings $\Delta x$, $\Delta y$.
The explicit scheme reads (Morton \& Mayers, Eq.~3.2):
\begin{equation}
U_{j,m}^{n+1} = U_{j,m}^n
  + \mu_x\bigl(U_{j+1,m}^n - 2U_{j,m}^n + U_{j-1,m}^n\bigr)
  + \mu_y\bigl(U_{j,m+1}^n - 2U_{j,m}^n + U_{j,m-1}^n\bigr)
\label{eq:app-ftcs-2d}
\end{equation}
where
\begin{equation}
\mu_x = \frac{\alpha\,\Delta t}{(\Delta x)^2},
\qquad
\mu_y = \frac{\alpha\,\Delta t}{(\Delta y)^2}\,.
\label{eq:app-mu-2d}
\end{equation}

\paragraph{Von Neumann ansatz.}
Substitute $U_{j,m}^n = \lambda^n\,e^{i(k_x j\Delta x + k_y m\Delta y)}$
into~\eqref{eq:app-ftcs-2d} and divide by the common exponential factor
(Eq.~3.8):
\begin{equation}
\boxed{\;\lambda(\mathbf{k})
  = 1
  - 4\mu_x\sin^2\!\frac{k_x\Delta x}{2}
  - 4\mu_y\sin^2\!\frac{k_y\Delta y}{2}\;}
\label{eq:app-amp-2d}
\end{equation}
This is Eq.~3.9 in Morton \& Mayers.

\paragraph{Stability condition.}
The worst case occurs when both $\sin^2$ terms equal 1
(the 2D Nyquist mode $k_x\Delta x = k_y\Delta y = \pi$).
Then $\lambda_{\min} = 1 - 4\mu_x - 4\mu_y$.
The condition $\lambda_{\min} \ge -1$ gives (Eq.~3.10):
\begin{equation}
\boxed{\;\mu_x + \mu_y \;\le\; \frac{1}{2}\;}
\label{eq:app-stability-2d}
\end{equation}
For equal spacing $\Delta x = \Delta y = h$, this becomes $2\mu \le 1/2$,
i.e.\ $\mu \le 1/4$ and $\Delta t \le h^2 / (4\,\alpha)$.

The maximum eigenvalue of the 2D discrete Laplacian (five-point stencil)
is
\begin{equation}
\lambda_{\max}^{\text{2D}} = \frac{4}{(\Delta x)^2} + \frac{4}{(\Delta y)^2}\,.
\label{eq:app-eigmax-2d}
\end{equation}
For the grid used in this work ($\Delta x = \Delta y = 0.5$),
$\lambda_{\max}^{\text{2D}} = 32$, and the unified stability
condition is $\alpha\,\Delta t\,\lambda_{\max}^{\text{2D}} < 2$,
consistent with Eq.~\eqref{eq:stability} in the main text.

\subsection{Application to the SAFE-PIT-CM grid}
\label{app:vn-application}

We now substitute the concrete grid parameters used throughout this work:
$\Delta x = \Delta y = 0.5$, $\Delta t_{\text{sim}} = 0.005$,
$\Delta t_{\text{save}} = 0.25$, $N_{\text{sub}} = 50$.

The maximum eigenvalue of the 2D five-point stencil is:
\begin{equation}
\lambda_{\max} = \frac{4}{(\Delta x)^2} + \frac{4}{(\Delta y)^2}
               = \frac{4}{0.25} + \frac{4}{0.25} = 16 + 16 = 32\,.
\label{eq:app-eigmax-grid}
\end{equation}

The Von Neumann stability number is
$\sigma = \Delta t_{\text{sub}} \cdot \alpha \cdot \lambda_{\max} < 2$.
Without sub-stepping ($N_{\text{sub}} = 1$,
$\Delta t_{\text{sub}} = \Delta t_{\text{save}} = 0.25$):
\begin{equation}
\sigma = 0.25 \times \alpha \times 32 = 8\,\alpha
\qquad\Rightarrow\qquad
\alpha_{\max} = \frac{2}{0.25 \times 32} = 0.25\,.
\label{eq:app-no-substep}
\end{equation}
Since the data spans $\alpha \in [0.01,\; 1.7]$, most simulations
violate the stability condition.

With sub-stepping ($N_{\text{sub}} = 50$,
$\Delta t_{\text{sub}} = 0.25 / 50 = 0.005$):
\begin{equation}
\sigma = 0.005 \times \alpha \times 32 = 0.16\,\alpha
\qquad\Rightarrow\qquad
\alpha_{\max} = \frac{2}{0.005 \times 32} = 12.5\,.
\label{eq:app-with-substep}
\end{equation}
This provides a $12.5 / 1.7 \approx 7\times$ safety margin over the
maximum diffusion coefficient in the data. Table~\ref{tab:vn-grid}
summarises both scenarios.

\begin{table}[h]
\centering
\caption{Von Neumann stability for the SAFE-PIT-CM grid
($\Delta x = \Delta y = 0.5$, $\lambda_{\max} = 32$).
The stability condition is $\sigma = \Delta t_{\text{sub}} \cdot \alpha \cdot \lambda_{\max} < 2$}
\label{tab:vn-grid}
\begin{tabular}{lcccc}
\toprule
\textbf{Scenario} & $\boldsymbol{\Delta t_{\text{sub}}}$ & $\boldsymbol{\sigma}$ \textbf{at} $\alpha = 1.7$ & $\boldsymbol{\alpha_{\max}}$ & \textbf{Stable?} \\
\midrule
No sub-stepping ($N_{\text{sub}} = 1$)  & $0.25$  & $13.6$ & $0.25$  & No  \\
Sub-stepping ($N_{\text{sub}} = 50$)    & $0.005$ & $0.27$ & $12.5$  & Yes \\
\bottomrule
\end{tabular}
\end{table}

\subsection{Stability (smoothing regime)}
\label{app:vn-sign}

For diffusion, the mesh Fourier number is $\mu = \alpha\,\Delta t / h^2$
with $\alpha > 0$.
The amplification factor $\lambda = 1 - 4\mu\,s \in [1 - 4\mu,\; 1]$ satisfies
$|\lambda| \le 1$ whenever $\mu \le 1/4$ (in 2D).
All Fourier modes \emph{decay}, and the scheme is \textbf{conditionally stable}.
This is the standard FTCS result. Because diffusion is a smoothing process,
the SAFE operator always runs in this stable regime.

\end{appendices}

\end{document}